\documentclass{article} 
\usepackage{nips15submit_e,times}
\usepackage{hyperref}
\usepackage{url}

\usepackage{amsmath}
\usepackage{amssymb}
\usepackage{dsfont}
\usepackage{booktabs,dcolumn}

\DeclareMathOperator*{\argmin}{arg\,min}

\usepackage{graphicx}
\usepackage{subfigure}

\usepackage{algorithm}
\usepackage{algorithmic}
\renewcommand{\algorithmicforall}{\textbf{for each}}

\title{A Variational Bayesian State-Space Approach to Online Passive-Aggressive Regression}

\author{
Arnold Salas\thanks{Corresponding author. This work was funded through AFR-PhD grant agreement 8837255 from the National Research Fund of Luxembourg, and by the Economic and Social Research Council (ESRC) and the Oxford-Man Institute.}\, ,\; Stephen J. Roberts and Michael A. Osborne\\
Department of Engineering Science and Oxford-Man Institute \\
University of Oxford \\
\texttt{arnold.salas@eng.ox.ac.uk},\; \texttt{\{sjrob, mosb\}@robots.ox.ac.uk}
}

\nipsfinalcopy 

\begin{document}

\maketitle

\begin{abstract}

Online Passive-Aggressive (PA) learning is a class of online margin-based algorithms suitable for a wide range of real-time prediction tasks, including classification and regression. PA algorithms are formulated in terms of deterministic point-estimation problems governed by a set of user-defined hyperparameters: the approach fails to capture model/prediction uncertainty and makes their performance highly sensitive to hyperparameter configurations. 
In this paper, we introduce a novel PA learning framework for regression that overcomes the above limitations. 
We contribute a Bayesian state-space interpretation of PA regression, along with a novel online variational inference scheme, that not only produces probabilistic predictions, but also offers the benefit of automatic hyperparameter tuning. Experiments with various real-world data sets show that our approach performs significantly better than a more standard, linear Gaussian state-space model. 

\end{abstract}

\section{Introduction}

Online learning is the most common approach of learning from non-stationary and/or large sequential data sets. In online learning, model parameters are learned in a sequential manner, thus achieving temporal adaptation and learning efficiency in time-aware applications. Among the popular algorithms, online Passive-Aggressive (PA) learning \cite{crammer06a} provides a generic family of online margin-based algorithms for various time-aware applications, including classification and regression. However, despite their merits, PA algorithms make point rather than probabilistic predictions, and depend on a set of hyperparameters that are assumed to be user-defined and constant over time. This assumption is impractical for at least two reasons. First, it has been recently argued that the performance of many machine learning algorithms is highly sensitive to hyperparameter settings \cite{hutter14}, and PA learning is unlikely to be an exception because its performance is measured in terms of cumulative loss. Second, in non-stationary environments, optimal hyperparameter choices may quickly become sub-optimal, due to the evolving nature of the underlying population distributions.

To address these drawbacks, we propose a new online PA method based on a Bayesian treatment of the existing PA framework. We concentrate here on PA learning for regression. Our algorithm incorporates a novel, online, variational inference scheme. Furthermore, it explicitly takes into account uncertainty in our predictions and is endowed with a self-tuning hyperparameter mechanism.

The main contributions of the paper are twofold. Firstly, this paper is, to the best of our knowledge, the first to approach online PA regression from a Bayesian state-space perspective. We will indeed show that the state-space representation of PA regression results in a Bayesian linear Gaussian state-space model (LGSSM). Secondly, we establish a clear connection between our online variational inference procedure and Streaming Variational Bayes \cite{svb}, thus making the first application of the latter to the Bayesian LGSSM setting.

\section{Bayesian State-Space Approach to Passive-Aggressive Regression}

In this section, we provide a Bayesian treatment of online PA regression within a state-space framework. We show that the state-space model (SSM) corresponding to PA regression is, conditionally upon the mean and variance of the measurement noise, a special case of the Bayesian LGSSM, and that it justifies the PA regression algorithm from a \emph{maximum a posteriori} (MAP) standpoint.

\subsection{Online Passive-Aggressive Regression}
\label{sec:par}

Consider a data stream consisting of examples $\left\{(\mathbf{x}_{t}, y_{t})\right\}_{t \geq 1}$, where $\mathbf{x}\in\mathbb{R}^{I}$ is an $I$-dimensional input vector and $y\in\mathbb{R}$ is the associated output. Online PA regression \cite{crammer06a} is based on the linear prediction model of the form $f(\mathbf{x}) = \mathbf{x}^{\top}\mathbf{w}$, where $\mathbf{w}\in\mathbb{R}^{I}$ is the incrementally learned weight vector. The PA regression algorithm initialises the weight vector to the zero vector ($\mathbf{\hat{w}}_{1} = \mathbf{0}_{I\times 1}$) and, after observing the $t^{\text{th}}$ example, the new weight $\mathbf{\hat{w}}_{t}$ is obtained as the solution to\footnote{We restrict our attention to the PA-I variant of PA regression.}
\begin{equation}
\label{eq:softpar-optprob}
	\min_{\mathbf{w}_{t}}\; \left\{\frac{1}{2}\left\Vert\mathbf{w}_{t} - \mathbf{\hat{w}}_{t-1}\right\Vert_{2}^{2} + C\ell\left(y_{t}, \mathbf{x}_{t}^{\top}\mathbf{w}_{t}; \epsilon\right)\right\},
\end{equation}
where $\ell(y, \hat{y}; \epsilon) = |y - \hat{y}|_{\epsilon} \equiv \max(|y - \hat{y}| - \epsilon, 0)$ is the $\epsilon$-insensitive loss function ($\epsilon$-ILF) and $C > 0$ is a user-specified parameter. The intuitive goal of PA regression is to minimally change the existing weight estimate while predicting the $t^{\text{th}}$ example as accurately as possible. The parameter $C$ serves to balance these two competing objectives. Larger values of $C$ imply a more aggressive update step, whence the name of \emph{aggressiveness parameter} \cite{crammer06a}.

\subsection{Bayesian Linear Gaussian State-Space Models}

LGSSMs\footnote{These are also called Kalman Filters/Smoothers and Linear Dynamical Systems.} are fundamental in time-series analysis \cite{grewal15, shumway10}. In these models, each output $y_{t}$ is generated from an underlying dynamical system on the hidden variable $\mathbf{h}_{t}$ according to:
\begin{equation}
\label{eq:bayes-lgsssm}
	y_{t}
	= \mathbf{b}^{\top}\mathbf{h}_{t} + \eta_{t},\quad		\eta_{t} \sim \mathcal{N}\left(\eta_{t}|0, \sigma^{2}\right),	\qquad
	\mathbf{h}_{t}
	= \mathbf{A}\mathbf{h}_{t-1} + \boldsymbol{\eta}_{t}^{\mathbf{h}},\quad		
	\boldsymbol{\eta}_{t}^{\mathbf{h}} \sim \mathcal{N}\left(\boldsymbol{\eta}_{t}^{\mathbf{h}}|\mathbf{0}_{H\times 1}, \boldsymbol{\Sigma}\right),
\end{equation}
where $H \equiv \dim(\mathbf{h}_{t})$. The initial latent variable also has a Gaussian distribution which we write as $p(\mathbf{h}_{1}) = \mathcal{N}(\mathbf{h}_{1}|\boldsymbol{\mu}_{\pi}, \boldsymbol{\Sigma}_{\pi})$. The model parameters are therefore $\boldsymbol{\theta} \equiv (\mathbf{A}, \mathbf{b}, \boldsymbol{\Sigma}, \sigma^{2}, \boldsymbol{\mu}_{\pi}, \boldsymbol{\Sigma}_{\pi})$. In the Bayesian treatment of the LGSSM, instead of considering $\boldsymbol{\theta}$ as fixed, we define a prior distribution $p(\boldsymbol{\theta}|\boldsymbol{\omega})$, where $\boldsymbol{\omega}$ is a vector of hyperparameters.

\subsection{Bayesian State-Space Representation of Passive-Aggressive Regression}

Let $\mathbf{I}_{I}$ be the identity matrix of order $I$. The state-space representation of PA regression is given by
\begin{equation}
\label{eq:par-ssm}
	y_{t} 
	= \mathbf{x}_{t}^{\top}\mathbf{w}_{t} + \eta_{t}, \quad	\eta_{t} \sim p\left(\eta_{t}|\epsilon\right),	
	\qquad
	\mathbf{w}_{t} 
	= \mathbf{w}_{t-1} + \boldsymbol{\eta}_{t}^{\mathbf{w}}, \quad		
		\boldsymbol{\eta}_{t}^{\mathbf{w}}
		\sim \mathcal{N}\left(\boldsymbol{\eta}_{t}^{\mathbf{w}}|\mathbf{0}_{I\times 1}, \alpha^{-1}\mathbf{I}_{I}\right),
\end{equation}
with the convention that $\mathbf{w}_{0} = \mathbf{0}_{I\times 1}$, and where
\begin{equation}
\label{eq:par-noisemodel}
	p\left(\eta_{t}|\epsilon\right)
	= \frac{1}{2(1+\epsilon)}e^{-\left|\eta_{t}\right|_{\epsilon}}
\end{equation}
is the measurement-noise density dictated by the $\epsilon$-ILF \cite{SmoSch04}. In this case, the weight posterior satisfies\footnote{For brevity, we omit $\mathbf{x}_{t}$ from the conditioning statements, and shall do so in the remainder of the paper.}
\begin{equation}
	p\left(\mathbf{w}_{t}|y_{t}, \mathbf{w}_{t-1}\right)
	\propto p\left(y_{t}|\mathbf{w}_{t}\right)p\left(\mathbf{w}_{t}|\mathbf{w}_{t-1}\right)
	= \frac{1}{2(1+\epsilon)}e^{-\ell(y_{t}, \mathbf{x}_{t}^{\top}\mathbf{w}_{t}; \epsilon)}
	  \mathcal{N}\left(\mathbf{w}_{t}|\mathbf{w}_{t-1}, \alpha^{-1}\mathbf{I}_{I}\right).
\end{equation}
Setting $(\alpha, \mathbf{w}_{t-1}) = (C^{-1}, \mathbf{\hat{w}}_{t-1})$ in the above equation, taking the negative logarithm thereof and ignoring any resulting additive constant yields the PA objective from \eqref{eq:softpar-optprob}. We thus obtain a MAP justification for the PA regression algorithm.

Observe that Eqs. \eqref{eq:par-ssm}-\eqref{eq:par-noisemodel} give a model that is intractable, due to the Laplacian-like noise distribution. Having said that, \cite{pontil2000} proved that this distribution can be expressed as a continuous mixture of Gaussians (CMoG). Specifically\footnote{We use a condensed integral notation: all integrals are definite integrals over the entire domain of interest.} ,
\begin{equation}
\label{eq:noisemodel-cMoG}
	p\left(\eta_{t}|\epsilon\right)
	= \int\int
		\mathcal{N}\left(\eta_{t}|\mu, \beta^{-1}\right)p\left(\mu|\epsilon\right)p\left(\beta\right)
	  \mathrm{d}\mu\mathrm{d}\beta,
\end{equation}
with
\begin{align}
\label{cMoG-priors}
	p\left(\beta\right)
	&= \mathcal{IG}\left(\beta|1, 1/2\right)
	= \frac{1}{2}\beta^{-2}e^{-\frac{1}{2\beta}}	\\
	p\left(\mu|\epsilon\right)
	&= \overline{\mathcal{U}}\left(\mu|-\epsilon, \epsilon\right)
	\equiv \frac{1}{2\left(1 + \epsilon\right)}\left[\mathds{1}_{\left[-\epsilon, \epsilon\right]}\left(\mu\right) 
	  	   + \delta\left(\mu + \epsilon\right) + \delta\left(\mu - \epsilon\right)\right],
\end{align}
where $\mathcal{IG}$ stands for `inverse Gamma', $\mathds{1}_{S}(\cdot)$ for the indicator function of the set $S$, and $\delta(\cdot)$ for the Dirac delta function. The above CMoG formulation implies that, conditionally upon $\beta$ and $\mu$, the SSM described by Eqs. \eqref{eq:par-ssm}-\eqref{eq:par-noisemodel} is a special case of the Bayesian LGSSM from \eqref{eq:bayes-lgsssm}. To retain this formalism, we will, in the first instance, hold $\beta$ and $\mu$ `fixed'. In the second instance, we will approximately marginalise $\beta$ and $\mu$ by means of an innovative, truly sequential, Variational Bayes (VB) routine.

Going forward, we shall refer to the ensuing model as \emph{BaYesian Passive-Aggressive State-Space Model}, or BYPASS for short. BYPASS additionally takes the prior over its parameter vector $\boldsymbol{\theta} = (\alpha, \beta, \mu)$ to factorise as
\begin{equation}
	p\left(\boldsymbol{\theta}|\boldsymbol{\omega}\right)
	= \mathcal{G}\left(\alpha|a, b\right)
	  \mathcal{IG}\left(\beta|1, 1/2\right)
	  \overline{\mathcal{U}}\left(\mu|-\epsilon, \epsilon\right),	\qquad
	\boldsymbol{\omega} 
	= \left(a, b, \epsilon\right).
\end{equation}
Note that we have assigned the standard conjugate prior to the weight precision $\alpha$. We do not define any prior for $\boldsymbol{\omega}$\footnote{A fully Bayesian treatment certainly requires the specification of a \emph{hyperprior}, but is not taken here for space restrictions.}. Probabilistically, the BYPASS model is defined by\footnote{$v_{1:t}$ denotes $v_{1}, \ldots, v_{t}$.}
\begin{equation}
	p\left(y_{1:t}, \mathbf{w}_{1:t}, \boldsymbol{\theta}|\boldsymbol{\omega}\right)
	= p\left(y_{1:t}, \mathbf{w}_{1:t}|\boldsymbol{\theta}\right)p\left(\boldsymbol{\theta}|\boldsymbol{\omega}\right)
	= \left[\prod_{\tau=1}^{t} p\left(y_{\tau}|\mathbf{w}_{\tau}, \mu, \beta\right)p\left(\mathbf{w}_{\tau}|\mathbf{w}_{\tau-1}, \alpha\right)\right]
	  p\left(\boldsymbol{\theta}|\boldsymbol{\omega}\right),
\end{equation}
where $p(y_{\tau}|\mathbf{w}_{\tau}, \mu, \beta) = \mathcal{N}(y_{\tau}|\mathbf{x}_{\tau}^{\top}\mathbf{w}_{\tau} + \mu, \beta^{-1})$ and $p(\mathbf{w}_{\tau}|\mathbf{w}_{\tau-1}, \alpha) = \mathcal{N}(\mathbf{w}_{\tau}|\mathbf{w}_{\tau-1}, \alpha^{-1}\mathbf{I}_{I})$.

\section{Genuinely Online Variational Inference}

An exact implementation of Bayesian LGSSMs is formally intractable \cite{davy03}. Besides sampling methods \cite{hmm_cappe, schnatter06}, VB approximations \cite{barber07, chiappa07} are popular approximate treatments in this context. Nonetheless, the drawback of such VB procedures is that they all require a full pass through the data at each iteration, rendering them impracticable for streaming data. To remedy this, we develop \emph{Genuinely Online Variational Inference} (GOVI), a novel framework whereby VB may be efficiently deployed in the streaming setting, without the need to revisit past data or have advance knowledge of future data. 
The rationale behind GOVI is to store the joint BYPASS distribution learned on round $t-1$ so as to recycle it in the subsequent round. This simple principle is reflected by the following probabilistic recursions:
\begin{align}
	p\left(y_{1:t-1}, \mathbf{w}_{1:t-1}|\langle \boldsymbol{\theta} \rangle_{1:t-1}\right)
	&= \prod_{\tau=1}^{t-1} p\left(y_{\tau}|\mathbf{w}_{\tau}, \langle \mu \rangle_{\tau}, \langle \beta \rangle_{\tau}\right)
	   p\left(\mathbf{w}_{\tau}|\mathbf{w}_{\tau-1}, \langle \alpha \rangle_{\tau}\right),	\label{eq:bypass-jointdist-recursion2} \\ 
	p\left(y_{1:t}, \mathbf{w}_{1:t}|\boldsymbol{\theta}, \langle \boldsymbol{\theta} \rangle_{1:t-1}\right)
	&= p\left(y_{t}|\mathbf{w}_{t}, \mu, \beta\right)p\left(\mathbf{w}_{t}|\mathbf{w}_{t-1}, \alpha\right)
	  p\left(y_{1:t-1}, \mathbf{w}_{1:t-1}|\langle \boldsymbol{\theta} \rangle_{1:t-1}\right),	\label{eq:bypass-jointdist-recursion1}
\end{align}
where $\langle \boldsymbol{\theta} \rangle_{t} \equiv \langle \boldsymbol{\theta} \rangle_{q_{t}(\boldsymbol{\theta})}$, $\langle \cdot \rangle_{d(x)}$ denotes the expectation w.r.t. the distribution $d(x)$, and $q_{t}(\cdot)$ is a shorthand for the approximating density $q(\cdot|y_{1:t}, \langle \boldsymbol{\theta} \rangle_{1:t-1})$. A crucial implication of this recycling process is that we may discard observations after processing them. As a result, GOVI is both single-pass and computationally efficient, thereby achieving the desiderata of streaming methods \cite{domingos03}.

To determine $q_{t}(\cdot)$, one considers the lower bound:
\begin{equation}
\label{eq:bypass-lowerbound}
	\log p\left(y_{1:t}|\langle \boldsymbol{\theta} \rangle_{1:t-1}, \boldsymbol{\omega}\right)
	\geq \langle E_{t}\left(\mathbf{w}_{1:t}, \boldsymbol{\theta}\right)\rangle_{q_{t}\left(\mathbf{w}_{1:t}, \boldsymbol{\theta}\right)}	
		    + \langle \log p\left(\boldsymbol{\theta}|\boldsymbol{\omega}\right)\rangle_{q_{t}\left(\boldsymbol{\theta}\right)}
		    + H\left(q_{t}\right)
	\equiv \mathcal{L},
\end{equation}
where $E_{t}(\mathbf{w}_{1:t}, \boldsymbol{\theta}) \equiv \log p(y_{1:t}, \mathbf{w}_{1:t}|\boldsymbol{\theta}, \langle \boldsymbol{\theta} \rangle_{1:t-1})$ and $H(d)$ signifies the entropy of $d(x)$. The key approximation in VB, commonly called the mean-field approximation (MFA), is $q_{t}(\mathbf{w}_{1:t}, \boldsymbol{\theta}) = q_{t}(\mathbf{w}_{1:t})\prod_{i}q_{t}(\theta_{i})$, from which one may show that, for optimality of $\mathcal{L}$,
\begin{equation}
	q_{t}\left(\mathbf{w}_{1:t}\right)
	\propto q(y_{1:t}, \mathbf{w}_{1:t}) 
	\equiv  e^{\langle E_{t}\left(\mathbf{w}_{1:t}, \boldsymbol{\theta}\right)\rangle_{q_{t}\left(\boldsymbol{\theta}\right)}},	\qquad
	q_{t}\left(\boldsymbol{\theta}\right)
	\propto p\left(\boldsymbol{\theta}|\boldsymbol{\omega}\right)
			e^{\langle E_{t}\left(\mathbf{w}_{1:t}, \boldsymbol{\theta}\right)\rangle_{q_{t}\left(\mathbf{w}_{1:t}\right)}}.	
\end{equation}
These coupled equations need to be iterated to convergence. Our main concern is with the update for $q_{t}(\mathbf{w}_{t})$, for which this paper makes a departure from treatments previously developed \cite{barber07, chiappa07}. We will present final results only, and refer the reader to the Supplementary Material for detailed derivations.

\subsection{Approximate Filtering}

From Eqs. \eqref{eq:bypass-jointdist-recursion2}-\eqref{eq:bypass-jointdist-recursion1}, it follows that
\begin{equation}
	q\left(y_{1:t}, \mathbf{w}_{1:t}\right)
	= \prod_{\tau=1}^{t}	\;
	 	\underbrace{\mathcal{N}\left(y_{\tau}|\mathbf{x}_{\tau}^{\top}\mathbf{w}_{\tau} + \langle \mu \rangle_{\tau}, \langle \beta \rangle_{\tau}^{-1}\right)}
	 		_{= q\left(y_{\tau}|\mathbf{w}_{\tau}\right) \approx p\left(y_{\tau}|\mathbf{w}_{\tau}, \mu, \beta\right)}	\;\,
		\times \, \underbrace{\mathcal{N}\left(\mathbf{w}_{\tau}|\mathbf{w}_{\tau-1}, \langle \alpha \rangle_{\tau}^{-1}\mathbf{I}_{I}\right)}
			_{= q\left(\mathbf{w}_{\tau}|\mathbf{w}_{\tau-1}\right) \approx p\left(\mathbf{w}_{\tau}|\mathbf{w}_{\tau-1}, \alpha\right)}.
\end{equation}
Clearly, the above represents the joint distribution of the BYPASS model with sequentially updated, averaged parameters. Thus, inference can be performed using the standard Kalman filter (KF) equations \cite{kalman60, zarchan09}. A direct consequence is that the approximate filtering distribution is Gaussian:
\begin{equation}
\label{eq:bypass-approx_filtering_dist}
	q_{t}\left(\mathbf{w}_{t}\right)
	= \mathcal{N}\left(\mathbf{w}_{t}|\boldsymbol{\mu}_{t}^{\mathbf{w}}, \boldsymbol{\Sigma}_{t}^{\mathbf{w}}\right).
\end{equation}
The moments of this distribution are iteratively updated as described in Algorithm \ref{alg:bypass}.

\subsection{Mean Variational Parameters}

\subsubsection*{Update for $\alpha$}

The approximate posterior over the weight precision is a Gamma distribution whose mean can be found from the following fixed-point iteration:
\begin{equation}
\label{eq:alpha_govi}
	\langle \alpha \rangle_{t}^{\text{new}}
	= \frac{2a}{2b + \left\Vert \boldsymbol{\mu}_{t}^{\mathbf{w}} - \boldsymbol{\mu}_{t-1}^{\mathbf{w}}\right\Vert_{2}^{2} 
				+ \mathrm{tr}\left(\boldsymbol{\Sigma}_{t}^{\mathbf{w}} - \boldsymbol{\Sigma}_{t-1}^{\mathbf{w}}\right)}.
\end{equation}

\subsubsection*{Update for $\beta$}

The variational posterior of $\beta$ is a generalised inverse Gaussian distribution defined by 
\begin{equation}
\label{eq:bypass-beta-vb_posterior}
	q_{t}\left(\beta\right)
	= \mathcal{GIG}\left(\beta|-1, 1, \rho_{t}\right),
	\qquad
	\rho_{t}
	= \left(y_{t} - \mathbf{x}_{t}^{\top}\boldsymbol{\mu}_{t}^{\mathbf{w}} - \langle \mu \rangle_{t}\right)^{2} 
	  + \mathbf{x}_{t}^{\top}\boldsymbol{\Sigma}_{t}^{\mathbf{w}}\mathbf{x}_{t} + \hat{V}_{t}^{\mu},
\end{equation}
where $\hat{V}_{t}^{\mu}$ denotes the variance of $\mu$ under $q_{t}$. The corresponding update equation is therefore
\begin{equation}
\label{eq:beta_govi}
	\langle \beta \rangle_{t}^{\text{new}}
	= \frac{\mathfrak{K}_{0}\left(\sqrt{\rho_{t}}\right)}{\sqrt{\rho_{t}}\mathfrak{K}_{1}\left(\sqrt{\rho_{t}}\right)},
\end{equation}
where $\mathfrak{K}_{\nu}(\cdot)$ denotes the modified Bessel function of the second kind, with index $\nu$.

\subsubsection*{Update for $\mu$}

The approximating density for $\mu$ is somewhat intractable and non-standard, but is roughly equal to a truncated Gaussian with lower and upper truncation values of $-\epsilon$ and $\epsilon$, respectively, so we set
\begin{equation}
	q_{t}\left(\mu\right)
	= \mathcal{N}^{\text{trunc}}_{\left[-\epsilon, \epsilon\right]}
	  \left(\mu|y_{t} - \mathbf{x}_{t}^{\top}\boldsymbol{\mu}_{t}^{\mathbf{w}}, \langle \beta \rangle_{t}^{-1}\right).
\end{equation}
From this, we obtain the following fixed-point equation in $\langle \mu \rangle_{t}$:
\begin{equation}
\label{eq:mu_govi}
	\langle \mu \rangle_{t}^{\text{new}}
	= y_{t} - \mathbf{x}_{t}^{\top}\boldsymbol{\mu}_{t}^{\mathbf{w}}
	  + \frac{\phi\left(l_{t}\right) - \phi\left(u_{t}\right)}
	  	{\sqrt{\langle \beta \rangle_{t}^{\text{old}}}\left[\Phi\left(u_{t}\right) - \Phi\left(l_{t}\right)\right]},
\end{equation}
where 
\begin{equation}
	l_{t} 
	= \sqrt{\langle \beta \rangle_{t}}\left[-\epsilon - \left(y_{t} - \mathbf{x}_{t}^{\top}\boldsymbol{\mu}_{t}^{\mathbf{w}}\right)\right],	\qquad
	u_{t} 
	= \sqrt{\langle \beta \rangle_{t}}\left[\epsilon - \left(y_{t} - \mathbf{x}_{t}^{\top}\boldsymbol{\mu}_{t}^{\mathbf{w}}\right)\right], 
\end{equation}
while $\phi(\cdot)$ and $\Phi(\cdot)$ denote the PDF and CDF of a standard Gaussian, respectively. Similarly,
\begin{equation}
\label{eq:varmu_govi}
	\left(\hat{V}_{t}^{\mu}\right)^{\text{new}}
	= \frac{1}{\langle \beta \rangle_{t}^{\text{old}}}
	  \left[1 + \frac{l_{t}\phi\left(l_{t}\right) - u_{t}\phi\left(u_{t}\right)}{\Phi\left(u_{t}\right) - \Phi\left(l_{t}\right)}
	        - \left(\frac{\phi\left(l_{t}\right) - \phi\left(u_{t}\right)}{\Phi\left(u_{t}\right) - \Phi\left(l_{t}\right)}\right)^{2}\right].
\end{equation}

\subsection{Relation to Streaming Variational Bayes}

In this section, we argue that GOVI falls under a broader family of online VB algorithms known as Streaming Variational Bayes (SVB) \cite{svb}. Note that Bayes' rule can be written in a streaming form:
\begin{equation}
	p\left(\Theta|y_{1:t}\right)
	\propto	p\left(y_{t}|\Theta\right)
			p\left(\Theta|y_{1:t-1}\right),
\end{equation}
where $\Theta$ represents a set of stochastic parameters. SVB suggests that, when the above is infeasible to compute, one should adopt an approximation algorithm $\mathcal{A}$ such that
\begin{equation}
	p\left(\Theta|y_{1:t}\right)
	\approx q_{t}\left(\Theta\right)
	= \mathcal{A}\left(y_{t}, q_{t-1}\left(\Theta\right)\right),
\end{equation}
with $q_{0}(\Theta) = p(\Theta)$. When $\mathcal{A}$ generates the posterior from Bayes' theorem, this calculation is exact.
In the setting of BYPASS, $\Theta = \left\{\mathbf{w}_{t}, \boldsymbol{\theta}\right\}$ and, by MFA, we obtain two separate approximation algorithms, namely
\begin{equation}
	q_{t}\left(\mathbf{w}_{t}\right)
	= \mathcal{A}_{\mathbf{w}}\left(y_{t}, q_{t-1}\left(\mathbf{w}_{t}\right)\right)
	\quad \text{and} \quad
	q_{t}\left(\boldsymbol{\theta}\right)
	= \mathcal{A}_{\boldsymbol{\theta}}\left(y_{t}, p\left(\boldsymbol{\theta}|\boldsymbol{\omega}\right)\right),
\end{equation}
the latter having to ineluctably rely on a time-invariant prior over $\boldsymbol{\theta}$, as the BYPASS framework does not specify any dynamics in that regard. More precisely, we have
\begin{align}
	q_{t}\left(\mathbf{w}_{t}\right)
	&\propto \mathcal{N}\left(y_{t}|\mathbf{x}_{t}^{\top}\mathbf{w}_{t} + \langle \mu \rangle_{t}, \langle \beta \rangle_{t}^{-1}\right)
			 \underbrace{\int \mathcal{N}\left(\mathbf{w}_{t}|\mathbf{w}_{t-1}, \langle \alpha \rangle_{t}^{-1}\mathbf{I}_{I}\right)
				q_{t-1}\left(\mathbf{w}_{t-1}\right) \mathrm{d}\mathbf{w}_{t-1}}_{= q_{t-1}\left(\mathbf{w}_{t}\right)}	\label{eq:bypass-svb_w}	\\
	q_{t}\left(\boldsymbol{\theta}\right)
	&\propto q_{t}\left(y_{t}|\boldsymbol{\theta}\right)p\left(\boldsymbol{\theta}|\boldsymbol{\omega}\right),	\quad
		q_{t}\left(y_{t}|\boldsymbol{\theta}\right)
		= \exp\bigg\{\langle \log p\left(y_{t}|\mathbf{w}_{t}, \mu, \beta\right)p\left(\mathbf{w}_{t}|\mathbf{w}_{t-1}, \alpha\right) \rangle_{q_{t}\left(\mathbf{w}_{1:t}\right)}\bigg\}.
\end{align}
Interestingly, from Eq. \eqref{eq:bypass-svb_w}, we are able to recover the KF equations evaluated at the mean variational parameters. The aforementioned digression from treatments previously presented thus emanates from the fact that we make the first application of SVB to the Bayesian LGSSM setting. 

\section{Learning the hyperparameters: Adaptive BYPASS}

As far as variational inference in Bayesian LGSSMs is concerned, the optimal hyperparameter values are typically obtained by optimising the variational lower bound $\mathcal{L}$ w.r.t. to $\boldsymbol{\omega}$ \cite{barber07, chiappa07}. However, this would not be computationally viable in a streaming environment. Since we are not treating $\boldsymbol{\omega}$ as a random vector, we may readily apply the PA regression framework from Section \ref{sec:par} to automatically tune $\boldsymbol{\omega}$ in an online manner. To mimic the ML-II (`evidence') framework, we use the negative log likelihood of the BYPASS model as the underlying loss function. This gives rise to the following optimisation problem:
\begin{equation}
\label{eq:adabypass-omega-opt}
	\boldsymbol{\hat{\omega}}_{t}
	= \argmin_{\boldsymbol{\omega} > \mathbf{0}_{M\times 1}}\; 
		\left\{\frac{1}{2}\left\Vert\boldsymbol{\omega} - \boldsymbol{\hat{\omega}}_{t-1}\right\Vert_{2}^{2} 
		+ C_{\boldsymbol{\omega}}\frac{\beta}{2}\left(y_{t} - \mathbf{x}_{t}^{\top}\boldsymbol{\mu}_{t-1}^{\mathbf{w}} - \mu\right)^{2}\right\},
\end{equation}
where $M \equiv \dim(\boldsymbol{\omega})$. We remark that, by construction, this problem corresponds to sequential maximum likelihood at the hyperparameter level. Its objective function depends on $\boldsymbol{\omega}$, insofar as the latter is employed to determine the weight estimates $\boldsymbol{\mu}_{t-1}^{\mathbf{w}}$. To convert this problem into a more `conventional' one, we replace the strict-positivity constraints $\boldsymbol{\omega} > \mathbf{0}_{M\times 1}$ by $\boldsymbol{\omega} \geq \boldsymbol{\omega}_{\text{min}}$, where $\boldsymbol{\omega}_{\text{min}} \approx \mathbf{0}_{M\times 1}$ represents a lower bound on $\boldsymbol{\omega}$. We consequently get (see Supplementary Material)
\begin{equation}
\label{eq:adabypass-omega-update}
	\boldsymbol{\hat{\omega}}_{t}
	= \max\bigg\{\boldsymbol{\hat{\omega}}_{t-1} 
	  + C_{\boldsymbol{\omega}}\langle \beta \rangle_{t-1}\mathbf{x}_{t}^{\top}\boldsymbol{\psi}_{t-1}\left(y_{t} - \mathbf{x}_{t}^{\top}\boldsymbol{\mu}_{t-1}^{\mathbf{w}} - \langle \mu \rangle_{t-1}\right)\mathbf{1}_{M\times 1},\; \boldsymbol{\omega}_{\text{min}}\bigg\},
\end{equation}
where the max operator is taken element-wise, $\mathbf{1}_{D\times 1}$ denotes a $D$-dimensional vector of ones and, for each $\omega\in\boldsymbol{\omega}$, 
$\boldsymbol{\psi}_{t}$ denotes the gradient of $\boldsymbol{\mu}_{t}^{\mathbf{w}}$ w.r.t. $\omega$ evaluated at $\omega = \hat{\omega}_{t}$.
As demonstrated in the Supplementary Material, this gradient is updated in an iterative fashion, based on its previous value and $\mathbf{S}_{t}$, the gradient of $\boldsymbol{\Sigma}_{t}^{\mathbf{w}}$ w.r.t. $\omega$ evaluated at $\hat{\omega}_{t}$. We dubbed the ensuing algorithm \emph{adaptive BYPASS} (ADA-BYPASS). The implementation details of the latter and of its non-adaptive counterpart are outlined in Algorithms \ref{alg:ada-bypass} and \ref{alg:bypass}, respectively.
\begin{algorithm}
   \caption{BYPASS}
   \label{alg:bypass}
\begin{algorithmic}[1]
   \STATE {\bfseries Input:} Hyperparameters $\boldsymbol{\omega}$, initial mean variational parameters $\langle \boldsymbol{\theta} \rangle_{0}$.
   \STATE Set $\boldsymbol{\mu}_{0}^{\mathbf{w}} = \mathbf{0}_{I\times 1}$ and $\boldsymbol{\Sigma}_{0}^{\mathbf{w}} = \mathbf{0}_{I\times I}$.
   \FOR{$t = 1, 2, \ldots$}
   \STATE Obtain new inputs $\mathbf{x}_{t}$.
   \STATE Compute the predictive mean and variance of the output:
   		  \begin{equation*}
   		  	\hat{m}_{t} = \mathbf{x}_{t}^{\top}\boldsymbol{\mu}_{t-1}^{\mathbf{w}} + \langle \mu \rangle_{t-1},	\qquad
   		  	\hat{V}_{t} = \mathbf{x}_{t}^{\top}\mathbf{P}_{t-1}^{\mathbf{w}}\mathbf{x}_{t} + \langle \beta \rangle_{t-1}^{-1}.
   		  \end{equation*}
   \STATE Derive the new mean variational parameters $\langle \boldsymbol{\theta} \rangle_{t}$ by repeating the fixed-point iterations \eqref{eq:alpha_govi}, \eqref{eq:beta_govi}, \eqref{eq:mu_govi} and \eqref{eq:varmu_govi} until convergence.
   \STATE Evaluate the \emph{predictive weight covariance} and the \emph{Kalman gain}:
   		  \begin{equation*}
   		  	\mathbf{P}_{t-1}^{\mathbf{w}}
			= \boldsymbol{\Sigma}_{t-1}^{\mathbf{w}} + \langle \alpha \rangle_{t}^{-1}\mathbf{I}_{I},	\qquad
			\mathbf{g}_{t}
			= \left(\mathbf{x}_{t}^{\top}\mathbf{P}_{t-1}^{\mathbf{w}}\mathbf{x}_{t} + \langle \beta \rangle_{t}^{-1}\right)^{-1}
	  		  \mathbf{P}_{t-1}^{\mathbf{w}}\mathbf{x}_{t}.
   		  \end{equation*}
   \STATE Update the mean and covariance of the approximate filtering distribution $q_{t}(\mathbf{w}_{t})$:
   		  \begin{equation*}
			\boldsymbol{\mu}_{t}^{\mathbf{w}}
			= \boldsymbol{\mu}_{t-1}^{\mathbf{w}} + \mathbf{g}_{t}\left(y_{t} - \mathbf{x}_{t}^{\top}\boldsymbol{\mu}_{t-1}^{\mathbf{w}} 
			  - \langle \mu \rangle_{t}\right),	\qquad
		  	\boldsymbol{\Sigma}_{t}^{\mathbf{w}}
		  	= \left(\mathbf{I}_{I} - \mathbf{g}_{t}\mathbf{x}_{t}^{\top}\right)\mathbf{P}_{t-1}^{\mathbf{w}}.
		  \end{equation*}
   \ENDFOR
\end{algorithmic}
\end{algorithm}
\begin{algorithm}[H]
   \caption{ADA-BYPASS: BYPASS with hyperparameter adaptation via PA regression.}
   \label{alg:ada-bypass}
\begin{algorithmic}[1]
   \STATE {\bfseries Input:} Initial hyperparameters $\boldsymbol{\hat{\omega}}_{0}$, lower hyperparameter bounds $\boldsymbol{\omega}_{\text{min}}$, initial mean variational parameters $\langle\boldsymbol{\theta}\rangle_{0}$, initial variational variance $\hat{V}_{0}^{\mu}$, aggressiveness parameter $C_{\boldsymbol{\omega}} > 0$.
   \STATE Same as Step 2 in Algorithm \ref{alg:bypass}. 
   \STATE Initialise the gradients w.r.t. $\omega\in\boldsymbol{\omega}$: $\boldsymbol{\psi}_{0} = \mathbf{0}_{I\times 1}$, $\mathbf{S}_{0} = \mathbf{I}_{I}$.
   \FOR{$t = 1, 2, \ldots$}
   \STATE Same as Steps 4-5 in Algorithm \ref{alg:bypass}.
   	\STATE Update the hyperparameters according to Eq. \eqref{eq:adabypass-omega-update}.
    \STATE Same as Steps 6-8 in Algorithm \ref{alg:bypass}.
    \STATE Update the gradients:
    	   \begin{align*}
    	   	\mathbf{S}_{t}
    	   	&= \left(\mathbf{I}_{I} - \mathbf{g}_{t}\mathbf{x}_{t}^{\top}\right)\mathbf{S}_{t-1}\left(\mathbf{I}_{I} - \mathbf{x}_{t}\mathbf{g}_{t}^{\top}\right),	\\
    	   	\boldsymbol{\psi}_{t}
    	   	&= \left(\mathbf{I}_{I} - \mathbf{g}_{t}\mathbf{x}_{t}^{\top}\right)\boldsymbol{\psi}_{t-1}
		 	   + \langle \beta \rangle_{t}\mathbf{S}_{t}\mathbf{x}_{t}
	  	         \left(y_{t} - \mathbf{x}_{t}^{\top}\boldsymbol{\mu}_{t-1}^{\mathbf{w}} - \langle \mu \rangle_{t}\right).
    	   \end{align*}
   \ENDFOR
\end{algorithmic}
\end{algorithm}

\section{Applications}

\subsection{Practicalities}

Based on the sensitivity analysis in \cite{crammer06a}, we set the aggressiveness parameter $C_{\boldsymbol{\omega}}$ equal to $10^{-3}$. The model parameters are initialised at their prior means, except for the output precision $\beta$, whose prior mean is undefined. A similar principle is applied to the variational variance of $\mu$. As a result of this, we obtain: $\langle\alpha\rangle_{0} = a/b$, $\langle\mu\rangle_{0} = 0$ and $\hat{V}_{0}^{\mu} = \mathrm{Var}\big[\,\overline{\mathcal{U}}(\mu|-\epsilon, \epsilon)\,\big] = \epsilon^{2}(1 + \epsilon/3) / (1 + \epsilon)$. As for $\beta$, we approximate its prior mean as follows: $\langle\beta\rangle_{0} \approx 0.5 / 10^{-3} = 500$.

Next, we choose initial values for the hyperparameters. In order to initially emulate the frequentist PA regression framework (Section \ref{sec:par}) while simultaneously making $p(\alpha|a, b)$ `uninformative' (i.e. broad), we set $a = C_{\boldsymbol{\omega}}^{-1}$ and $b = 1$. As for the insensitivity hyperparameter, we use $\epsilon = 1.25$, this value being the mean of a symmetric Beta distribution of the second kind\footnote{We show in the Supplementary Material that the form of $p(\eta_{t}|\epsilon)$ induces this prior for $\epsilon$.} with shape parameter $s = 5$, the choice of which was motivated by \cite{vb_gar}. Finally, we selected $\omega_{\text{min}} = 10^{-8},\, \forall \omega_{\text{min}}\in\boldsymbol{\omega}_{\text{min}}$.

\subsection{Model specification and benchmark}

In the following experiments, unless otherwise stated, we used an autoregressive measurement equation of order 1 (AR(1)): $y_{t} = w_{t, 0} + w_{t, 1}y_{t-1} + \eta_{t}$, where $w_{t, 0}$ is a bias parameter. While this is perhaps not the best specification, feature selection goes beyond the scope of the present study. It is worthwhile noting, however, that there is no theoretical or practical obstacle that would prevent us from considering more complex predictors. This would be expected to further improve the model's performance.

We make comparisons with a standard LGSSM in which a MAP recursion is used to govern the adaptation of the model parameters, by sequentially using the maximum-likelihood formulation first proposed by \cite{jazwinski70}. To ensure full comparability of results, we also endow this model with an AR(1) hypothesis, and refer to it as \emph{sequential Kalman filter} (SKF) in the applications below.

In both models, one-step ahead forecasts are successively iterated to provide multi-step forecasts of arbitrary length, as needed. Missing values, if they occur, are accommodated for using the scheme advocated by \cite{shumway10}, in which they are replaced by their expectations under the corresponding model.

\subsection{Nile data}

We first consider a canonical changepoint data set, the minimum water levels of the Nile river during the period AD $622$-$1284$ \cite{whitcher02}. Several authors have found evidence supporting a changepoint for these data around AD $720$-$722$ \cite{whitcher02, garnett10, ray02}. The conjectured reason for this changepoint is the construction in AD 715 of a new device (a `nilometer') on the island of Roda, which affected the nature and accuracy of the measurements.

\begin{figure}[!htb]
 \begin{minipage}[c]{0.5\linewidth}
    \includegraphics[scale=0.5]{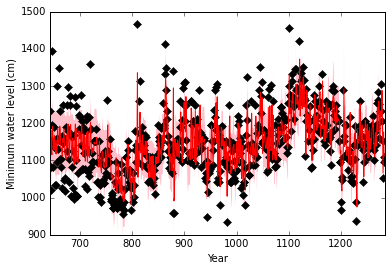}
 \end{minipage}
 \quad
 \begin{minipage}[c]{0.45\linewidth}
 	\begin{table}[H]
 	\begin{tabular}{lcc}
	\toprule
	{Metric}  & {ADA-BYPASS} & {SKF} \\
	\midrule
	RMSE (cm)        &$\mathbf{0.72}$				   &$0.85$	\\
	MAD (cm)         &$\mathbf{0.42}$ 				   &$0.45$ 	\\
	MAE (cm)		 &$\mathbf{0.54}$				   &$0.62$	\\
	LL   			 &$\mathbf{-754.2}$ 			   &$-971.78$	\\
	\bottomrule
	\end{tabular}
	\end{table}
 \end{minipage}
 \caption{Online one-year ahead predictions for the Nile's minimum water levels. Left panel: observed levels (black diamonds), predicted levels (red line) and $\pm 1$ standard deviation error bars (pink area). Right panel: predictive performances; error metrics shown are root mean squared error (RMSE), mean absolute deviation (MAD), mean absolute error (MAE) and predictive log likelihood (LL).}
 \label{fig:nile}
\end{figure}

We performed one-year lookahead prediction on this data set. The results can be seen in Fig. \ref{fig:nile}. We note the superior performance of ADA-BYPASS compared with the SKF. 

\subsection{Wind speed data}

To demonstrate the superior performance of ADA-BYPASS on a large data set, we next present the series of anemometer wind speed measurements (in m/s) from a Danish wind turbine. The data were sampled at 10 minute intervals for just over nine months, resulting in a total of 40,174 measurements. The 10 minute lookahead predictive performance achieved by each method is reported in Table \ref{table:ws}.
\begin{table}[h]
 	\caption{Predictive performance of ADA-BYPASS vs SKF on the wind speed data set.}
 	\label{table:ws}
 	\begin{center}
 	\begin{tabular}{lcc}
	\toprule
	{Metric}  & {ADA-BYPASS} & {SKF} \\
	\midrule
	RMSE (m/s)  &$\mathbf{0.6}$			   &$0.64$	\\
	MAD (m/s)   &$\mathbf{0.3}$ 			   &$0.31$ 	\\
	MAE (m/s)	  &$\mathbf{0.42}$			   &$0.44$	\\
	LL   		  &$\mathbf{-24,971.75}$	   &$-30,140.04$	\\
	\bottomrule
 	\end{tabular}
 	\end{center}
\end{table}

\subsection{Statistical Arbitrage}

LGSSMs, and variants thereof, have seen a widespread use in statistical arbitrage strategies, notably in pairs trading \cite{epchan2, montana11, montana09}. In this area, they serve as a dynamic model for the price spread between two assets. In our application, we seek to find the \emph{hedge ratio}\footnote{The hedge ratio of a particular asset is the number of units of that asset we should buy or sell in a portfolio. If the asset is a stock, then the number of units corresponds to the number of shares. A negative hedge ratio indicates we should sell that asset.} and the predictive standard deviation of the spread. The observable variable is thus one of the price series $y$, and the hidden variable is the hedge ratio $w$. We assume that both variables obey the ADA-BYPASS dynamics, i.e.
\begin{equation}
	y_{t}
	= w_{t}x_{t} + \eta_{t},	\quad
	\eta_{t}
	\sim \mathcal{N}\left(\eta_{t}|\mu, \beta^{-1}\right),	\qquad
	w_{t} 
	= w_{t-1} + \zeta_{t},	\quad
	\zeta_{t} 
	\sim \mathcal{N}\left(\zeta_{t}|0, \alpha^{-1}\right),
\end{equation}
where $x$ is the price series of the other asset. Typically, $\alpha,\,\beta$ and $\mu$ are manually selected in hindsight \cite{epchan2}. However, this practice is highly prone to the so-called \emph{data-snooping bias}: these parameters can be tweaked so as to optimise the backtesting performance of the strategy. The ADA-BYPASS algorithm automatically tunes its underlying parameters, so it does not suffer from this caveat.

We tested ADA-BYPASS on a pair of exchange-traded funds (ETFs) consisting of the SPDR gold trust GLD and the gold-miners ETF GDX. This ETF pairing is a favourite in the financial industry, because the value of gold-mining companies is very much based on the value of gold. We downloaded the corresponding, daily adjusted closing prices from Yahoo! Finance, between 22/05/2006 and 22/04/2015. 

Rather than maximising profits, most investors attempt to maximise risk-adjusted returns, as advocated by modern portfolio theory. The Sharpe ratio is the most widely used measure of risk-adjusted returns \cite{sharpe66}. Besides the Sharpe ratio, the maximum drawdown and maximum drawdown duration are two other popular metrics to evaluate trading strategies. From Table \ref{table:gdx-gld}, we can clearly discern that ADA-BYPASS beats SKF by a significant margin in terms of the aforementioned performance metrics.
\begin{table}[h]
 	\caption{Performance of the GDX-GLD pairs trade under ADA-BYPASS and SKF.}
 	\label{table:gdx-gld}
 	\begin{center}
 	\begin{tabular}{lcc}
	\toprule
	{Metric}  & {ADA-BYPASS} & {SKF} \\
	\midrule
	Sharpe ratio  &$\mathbf{1.12}$			   &$0.7$	\\
	Maximum drawdown (\%)    &$\mathbf{14.61}$ 			   &$73.05$ 	\\
	Maximum drawdown duration (trading days)	  &$\mathbf{375}$			   &$567$	\\
	\bottomrule
 	\end{tabular}
 	\end{center}
\end{table}

\section{Concluding remarks}

We introduced the first online Bayesian PA regression model within the state-space setting, along with a novel, online variational inference algorithm. This model is ideal for the probabilistic prediction of non-stationary and/or very large time series, in particular massive, time-varying data streams. Results on three real-world data sets show significant improvements in predictive performance over a more standard LGSSM.

\subsubsection*{References}

\renewcommand\refname{\vskip -0.9cm}

\small
{
\bibliographystyle{unsrt}
\bibliography{bypass}
}


\end{document}


\maketitle

\section*{Genuinely Online Variational Inference}

\subsubsection*{Update for $\alpha$}

The approximate posterior over the weight precision can be written as
\begin{equation}
	q_{t}\left(\alpha\right)
	\propto \mathcal{G}\left(\alpha|a, b\right)
			e^{\langle E_{t}\left(\mathbf{w}_{1:t}, \boldsymbol{\theta}\right)\rangle_{q_{t}\left(\mathbf{w}_{1:t}, \mu, \beta\right)}},
\end{equation}
where $E_{t}(\mathbf{w}_{1:t}, \boldsymbol{\theta}) \equiv \log p(y_{1:t}, \mathbf{w}_{1:t}|\boldsymbol{\theta}, \langle \boldsymbol{\theta} \rangle_{1:t-1})$. We have
\begin{equation}
\begin{split}
	E_{t}\left(\mathbf{w}_{1:t}, \boldsymbol{\theta}\right)
	&= \log p\left(y_{t}|\mathbf{w}_{t}, \mu, \beta\right) + \log p\left(\mathbf{w}_{t}|\mathbf{w}_{t-1}, \alpha\right) + \log p\left(y_{1:t-1}, \mathbf{w}_{1:t-1}|\langle \boldsymbol{\theta} \rangle_{1:t-1}\right) \\
	&= -\frac{\beta}{2}\left(y_{t} - \mathbf{x}_{t}^{\top}\mathbf{w}_{t} - \mu\right)^{2} - \frac{\alpha}{2}\left\Vert \mathbf{w}_{t} - \mathbf{w}_{t-1}\right\Vert_{2}^{2}	\\
	&\quad	 -\frac{1}{2}\sum_{\tau=1}^{t-1} \left[\langle \beta \rangle_{\tau}\left(y_{\tau} - \mathbf{x}_{\tau}^{\top}\mathbf{w}_{\tau} - \langle \mu \rangle_{\tau}\right)^{2} 
		 			 + \langle \alpha \rangle_{\tau}\left\Vert \mathbf{w}_{\tau} - \mathbf{w}_{\tau-1}\right\Vert_{2}^{2}\right] + \mathcal{Z},
\end{split}
\end{equation}
where $\mathcal{Z}$ regroups the normalising constants. Hence,
\begin{equation}
	\langle E_{t}\left(\mathbf{w}_{1:t}, \boldsymbol{\theta}\right)\rangle_{q_{t}\left(\mathbf{w}_{1:t}, \mu, \beta\right)}
	= -\frac{\alpha}{2}\langle \left\Vert \mathbf{w}_{t} - \mathbf{w}_{t-1}\right\Vert_{2}^{2}\rangle + \text{const},
\end{equation}
where the constant term does not depend on $\alpha$ and we have used $\langle\left\Vert \mathbf{w}_{t} - \mathbf{w}_{t-1}\right\Vert_{2}^{2}\rangle$ as a shorthand for $\langle\left\Vert \mathbf{w}_{t} - \mathbf{w}_{t-1}\right\Vert_{2}^{2}\rangle_{q_{t}(\mathbf{w}_{1:t})}$. Since $q_{t}(\mathbf{w}_{t}) = \mathcal{N}(\mathbf{w}_{t}|\boldsymbol{\mu}_{t}^{\mathbf{w}}, \boldsymbol{\Sigma}_{t}^{\mathbf{w}})$, the difference $\mathbf{w}_{t} - \mathbf{w}_{t-1}$ follows a Gaussian distribution with mean $\boldsymbol{\mu}_{t}^{\mathbf{w}} - \boldsymbol{\mu}_{t-1}^{\mathbf{w}}$ and covariance $\boldsymbol{\Sigma}_{t}^{\mathbf{w}} - \boldsymbol{\Sigma}_{t-1}^{\mathbf{w}}$, and so
\begin{equation}
	\langle\left\Vert \mathbf{w}_{t} - \mathbf{w}_{t-1}\right\Vert_{2}^{2}\rangle
	= \left\Vert \boldsymbol{\mu}_{t}^{\mathbf{w}} - \boldsymbol{\mu}_{t-1}^{\mathbf{w}}\right\Vert_{2}^{2} 
		+ \mathrm{tr}\left(\boldsymbol{\Sigma}_{t}^{\mathbf{w}} - \boldsymbol{\Sigma}_{t-1}^{\mathbf{w}}\right).
\end{equation}
We deduce that
\begin{equation}
	q_{t}\left(\alpha\right)
	= \mathcal{G}\left(\alpha|a,\; b + \langle\left\Vert \mathbf{w}_{t} - \mathbf{w}_{t-1}\right\Vert_{2}^{2}\rangle/2\right),
\end{equation}
which gives rise to the re-estimation rule
\begin{equation}
\label{eq:bypass-alphahat-em_update}
	\langle \alpha \rangle_{t}^{\text{new}}
	= \frac{2a}{\langle\left\Vert \mathbf{w}_{t} - \mathbf{w}_{t-1}\right\Vert_{2}^{2}\rangle + 2b}.
\end{equation}
Interestingly, this rule is reminiscent of the corresponding expectation-maximisation update in the context of Bayesian regression \cite{mackay92a, tipping01}.

\subsection*{Update for $\beta$}

The variational posterior for the precision of the output satisfies
\begin{equation}
	q_{t}\left(\beta\right)
	\propto \mathcal{IG}\left(\beta|1, 1/2\right)
			e^{\langle E_{t}\left(\mathbf{w}_{1:t}, \boldsymbol{\theta}\right)\rangle_{q_{t}\left(\mathbf{w}_{1:t}, \alpha, \mu\right)}}.
\end{equation}
Up to an additive term that is independent of $\beta$, the exponent in the above equation can be rewritten as
\begin{equation}
\begin{split}
	\langle E_{t}\left(\mathbf{w}_{1:t}, \boldsymbol{\theta}\right)\rangle_{q_{t}\left(\mathbf{w}_{1:t}, \alpha, \mu\right)}
	&= -\frac{\beta}{2}\big\langle\left(y_{t} - \mathbf{x}_{t}^{\top}\mathbf{w}_{t} - \mu\right)^{2} 
		\big\rangle_{q_{t}\left(\mu\right)q_{t}\left(\mathbf{w}_{t}\right)}	\\
	&= -\frac{\beta}{2}\big\langle\left(y_{t} - \mathbf{x}_{t}^{\top}\mathbf{w}_{t} - \langle \mu \rangle_{t}\right)^{2} 
	   + \hat{V}_{t}^{\mu}\big\rangle_{q_{t}\left(\mathbf{w}_{t}\right)}	\\
	&= -\frac{\beta}{2}\left[\left(y_{t} - \mathbf{x}_{t}^{\top}\boldsymbol{\mu}_{t}^{\mathbf{w}} - \langle \mu \rangle_{t}\right)^{2} 
	   + \mathbf{x}_{t}^{\top}\boldsymbol{\Sigma}_{t}^{\mathbf{w}}\mathbf{x}_{t} + \hat{V}_{t}^{\mu}\right].
\end{split}
\end{equation}
As a result, we obtain
\begin{equation}
	q_{t}\left(\beta\right)
	= \mathcal{GIG}\left(\beta|-1, 1, \rho_{t}\right),
\end{equation}
where
\begin{equation}
	\rho_{t} 
	\equiv \left(y_{t} - \mathbf{x}_{t}^{\top}\boldsymbol{\mu}_{t}^{\mathbf{w}} - \langle \mu \rangle_{t}\right)^{2} 
	  + \mathbf{x}_{t}^{\top}\boldsymbol{\Sigma}_{t}^{\mathbf{w}}\mathbf{x}_{t} + \hat{V}_{t}^{\mu}
\end{equation}
and
\begin{equation}
	\mathcal{GIG}\left(r|\nu, \chi, \rho\right)
	= \frac{\left(\rho/\chi\right)^{\nu/2}}{2\mathfrak{K}_{\nu}\left(\sqrt{\chi\rho}\right)}
	  r^{\nu-1}e^{-\left(\chi r^{-1} + \rho r\right)/2}	\qquad \left(r > 0\right)
\end{equation}
is the density of the Generalised Inverse Gaussian distribution \cite{jorgensen82}. The term $\mathfrak{K}_{\nu}(\cdot)$ represents the modified Bessel function of the second kind and with index $\nu$ \cite{abramowitz+stegun, watson44}. 
The variational mean of $\beta$ is therefore given by
\begin{equation}
	\langle \beta \rangle_{t}
	= \frac{\mathfrak{K}_{0}\left(\sqrt{\rho_{t}}\right)}{\sqrt{\rho_{t}}\mathfrak{K}_{1}\left(\sqrt{\rho_{t}}\right)}.
\end{equation}

\subsection*{Update for $\mu$}

The approximating posterior density of $\mu$ is defined by
\begin{equation}
	q_{t}\left(\mu\right)
	\propto \overline{\mathcal{U}}\left(\mu|-\epsilon, \epsilon\right)
			e^{\langle E_{t}\left(\mathbf{w}_{1:t}, \boldsymbol{\theta}\right)\rangle_{q_{t}\left(\mathbf{w}_{1:t}, \alpha, \beta\right)}},
\end{equation}
with
\begin{equation}
\begin{split}
	\langle E_{t}\left(\mathbf{w}_{1:t}, \boldsymbol{\theta}\right)\rangle_{q_{t}\left(\mathbf{w}_{1:t}, \alpha, \beta\right)}
	&= -\frac{\langle \beta \rangle_{t}}{2}
	  \big\langle \left(y_{t} - \mathbf{x}_{t}^{\top}\mathbf{w}_{t} - \mu\right)^{2} \big\rangle_{q_{t}\left(\mathbf{w}_{t}\right)}
	  + \text{const}	\\
	&= -\frac{\langle \beta \rangle_{t}}{2}\left[\mu - \left(y_{t} - \mathbf{x}_{t}^{\top}\boldsymbol{\mu}_{t}^{\mathbf{w}}\right)\right]^{2}
	  + \text{const}.
\end{split}
\end{equation}
It immediately follows that 
\begin{equation}
	q_{t}\left(\mu\right)
	\propto \overline{\mathcal{U}}\left(\mu|-\epsilon, \epsilon\right)
			\mathcal{N}\left(\mu|y_{t} - \mathbf{x}_{t}^{\top}\boldsymbol{\mu}_{t}^{\mathbf{w}}, \langle \beta \rangle_{t}^{-1}\right).
\end{equation}
Although this is an unusual distribution, it may be well approximated by a truncated Gaussian with lower and upper limits equal to $-\epsilon$ and $\epsilon$, respectively. That is,
\begin{equation}
\label{eq:bypass_mu_approxdist}
\begin{split}
	q_{t}\left(\mu\right)
	\approx \tilde{q}_{t}\left(\mu\right)
	&= \mathcal{N}^{\text{trunc}}_{\left[-\epsilon, \epsilon\right]}
	   \left(\mu|y_{t} - \mathbf{x}_{t}^{\top}\boldsymbol{\mu}_{t}^{\mathbf{w}}, \langle \beta \rangle_{t}^{-1}\right)	\\
	&= \mathds{1}_{\left[-\epsilon, \epsilon\right]}\left(\mu\right)
	   \frac{\sqrt{\langle \beta \rangle_{t}}\phi\left(\sqrt{\langle \beta \rangle_{t}}\left[\mu - \left(y_{t} - \mathbf{x}_{t}^{\top}\boldsymbol{\mu}_{t}^{\mathbf{w}}\right)\right]\right)}{\Phi\left(u_{t}\right) - \Phi\left(l_{t}\right)},
\end{split}
\end{equation}
where the normalised lower and upper bounds $l_{t}$ and $u_{t}$ are defined as
\begin{equation}
	l_{t} 
	= \sqrt{\langle \beta \rangle_{t}}\left[-\epsilon - \left(y_{t} - \mathbf{x}_{t}^{\top}\boldsymbol{\mu}_{t}^{\mathbf{w}}\right)\right],	\qquad
	u_{t} 
	= \sqrt{\langle \beta \rangle_{t}}\left[\epsilon - \left(y_{t} - \mathbf{x}_{t}^{\top}\boldsymbol{\mu}_{t}^{\mathbf{w}}\right)\right], 
\end{equation}
while $\phi(\cdot)$ and $\Phi(\cdot)$ are the PDF and CDF of the standard Normal distribution, respectively.

Given the above approximation, we redefine $\langle \mu \rangle_{t}$ as $\langle \mu \rangle_{\tilde{q}_{t}(\mu)}$, and similarly for the corresponding variational variance $\hat{V}_{t}^{\mu}$. This leads to the following fixed-point iterations \cite{johnson94}:
\begin{align}
	\langle \mu \rangle_{t}^{\text{new}}
	&= y_{t} - \mathbf{x}_{t}^{\top}\boldsymbol{\mu}_{t}^{\mathbf{w}}
	  + \frac{\phi\left(l_{t}\right) - \phi\left(u_{t}\right)}
	  	{\sqrt{\langle \beta \rangle_{t}^{\text{old}}}\left[\Phi\left(u_{t}\right) - \Phi\left(l_{t}\right)\right]},	\\
	\left(\hat{V}_{t}^{\mu}\right)^{\text{new}}
	&= \frac{1}{\langle \beta \rangle_{t}^{\text{old}}}
	  \left[1 + \frac{l_{t}\phi\left(l_{t}\right) - u_{t}\phi\left(u_{t}\right)}{\Phi\left(u_{t}\right) - \Phi\left(l_{t}\right)}
	        - \left(\frac{\phi\left(l_{t}\right) - \phi\left(u_{t}\right)}{\Phi\left(u_{t}\right) - \Phi\left(l_{t}\right)}\right)^{2}\right].
\end{align}

\section*{Learning the hyperparameters: Adaptive BYPASS}

\subsection*{Hyperparameter adaptation}

The PA hyperparameter updates are given by
\begin{equation}
\label{eq:adabypass-omega-opt}
	\boldsymbol{\hat{\omega}}_{t}
	= \argmin_{\boldsymbol{\omega} \geq \boldsymbol{\omega}_{\text{min}}}\; 
		\left\{\frac{1}{2}\left\Vert\boldsymbol{\omega} - \boldsymbol{\hat{\omega}}_{t-1}\right\Vert_{2}^{2} 
		+ C_{\boldsymbol{\omega}}\frac{\beta}{2}\left(y_{t} - \mathbf{x}_{t}^{\top}\boldsymbol{\mu}_{t-1}^{\mathbf{w}} - \mu\right)^{2}\right\}.
\end{equation}
This optimisation problem has a convex objective function and feasible affine constraints. These are sufficient conditions for Slater's condition to hold. Therefore, satisfying the Karush-Kuhn-Tucker (KKT) conditions is a necessary and sufficient condition for optimality \cite{BoydVand04}. The corresponding Lagrangian is
\begin{equation}
	\mathfrak{L}\left(\boldsymbol{\omega}, \boldsymbol{\lambda}\right)
	= \frac{1}{2}\left\Vert\boldsymbol{\omega} - \boldsymbol{\hat{\omega}}_{t-1}\right\Vert_{2}^{2} 
	  + C_{\boldsymbol{\omega}}\frac{\beta}{2}\left(y_{t} - \mathbf{x}_{t}^{\top}\boldsymbol{\mu}_{t-1}^{\mathbf{w}} - \mu\right)^{2}
	  + \boldsymbol{\lambda}^{\top}\left(\boldsymbol{\omega}_{\text{min}} - \boldsymbol{\omega}\right),
\end{equation}
where $\boldsymbol{\lambda}\in\mathbb{R}_{+}^{M}$ are Lagrange multipliers. Differentiating the Lagrangian w.r.t. $\boldsymbol{\omega}$ and solving for zero gives
\begin{equation}
	\boldsymbol{\omega}
	= \boldsymbol{\hat{\omega}}_{t-1} 
	  + C_{\boldsymbol{\omega}}\,\beta\mathbf{x}_{t}^{\top}\boldsymbol{\psi}_{t-1}\left(y_{t} - \mathbf{x}_{t}^{\top}\boldsymbol{\mu}_{t-1}^{\mathbf{w}} 
	    - \mu\right)\mathbf{1}_{M\times 1} + \boldsymbol{\lambda}.
\end{equation}
The KKT complementary slackness conditions require
that $(\omega_{j} - \omega_{\text{min}, j})\lambda_{j} = 0,\,\forall j$. It follows that if $\omega_{j} > \omega_{\text{min}, j}$, then $\lambda_{j} = 0$. Otherwise, the hyperparameter constraints imply that $\omega_{j} = \omega_{\text{min}, j}$. Wrapping up the two cases, we obtain the following update:
\begin{equation}
\label{eq:adabypass-omega-update}
	\boldsymbol{\omega}
	= \max\bigg\{\boldsymbol{\hat{\omega}}_{t-1}
	  + C_{\boldsymbol{\omega}}\,\beta\mathbf{x}_{t}^{\top}\boldsymbol{\psi}_{t-1}\left(y_{t} - \mathbf{x}_{t}^{\top}\boldsymbol{\mu}_{t-1}^{\mathbf{w}} 
	    - \mu\right)\mathbf{1}_{M\times 1},\; \boldsymbol{\omega}_{\text{min}}\bigg\}.
\end{equation}
Finally, replacing $\beta$ and $\mu$ with their respective variational means at $t-1$ yields $\boldsymbol{\hat{\omega}}_{t}$.

\subsection*{Hyperparameter gradients}

This section is largely inspired by \cite{haykin13}. Let $\omega\in\boldsymbol{\omega}$ be any hyperparameter and define
\begin{equation}
	\boldsymbol{\psi}_{t}
	\equiv \frac{\partial \boldsymbol{\mu}_{t}^{\mathbf{w}}}{\partial\omega},	\qquad
	\mathbf{S}_{t}
	\equiv \frac{\partial \boldsymbol{\Sigma}_{t}^{\mathbf{w}}}{\partial\omega}.
\end{equation}
Recall that
\begin{equation}
	\boldsymbol{\mu}_{t}^{\mathbf{w}}
	= \boldsymbol{\mu}_{t-1}^{\mathbf{w}} 
	  + \mathbf{g}_{t}\left(y_{t} - \mathbf{x}_{t}^{\top}\boldsymbol{\mu}_{t-1}^{\mathbf{w}} - \langle \mu \rangle_{t}\right),
\end{equation}
where
\begin{equation}
	\mathbf{g}_{t}
	= \left(\mathbf{x}_{t}^{\top}\mathbf{P}_{t-1}^{\mathbf{w}}\mathbf{x}_{t} + \langle \beta \rangle_{t}^{-1}\right)^{-1}
	  		  \mathbf{P}_{t-1}^{\mathbf{w}}\mathbf{x}_{t}
\end{equation}
is the Kalman gain. By rearranging this equation, we have
\begin{equation}
	\langle \beta \rangle_{t}^{-1}\mathbf{g}_{t}
	= \left(\mathbf{P}_{t-1}^{\mathbf{w}} - \mathbf{g}_{t}\mathbf{x}_{t}^{\top}\mathbf{P}_{t-1}^{\mathbf{w}}\right)\mathbf{x}_{t}.
\end{equation}
The expression inside the brackets on the RHS of this equation equals $\boldsymbol{\Sigma}_{t}^{\mathbf{w}}$. Hence, we may simplify the Kalman gain to
\begin{equation}
	\mathbf{g}_{t}
	= \langle \beta \rangle_{t}\boldsymbol{\Sigma}_{t}^{\mathbf{w}}\mathbf{x}_{t}.
\end{equation}
It immediately follows that
\begin{equation}
\begin{split}
	\boldsymbol{\psi}_{t}
	&= \boldsymbol{\psi}_{t-1} - \mathbf{g}_{t}\mathbf{x}_{t}^{\top}\boldsymbol{\psi}_{t-1} 
	   + \langle \beta \rangle_{t}\mathbf{S}_{t}\mathbf{x}_{t}
	  	 \left(y_{t} - \mathbf{x}_{t}^{\top}\boldsymbol{\mu}_{t-1}^{\mathbf{w}} - \langle \mu \rangle_{t}\right)	\\
	&= \left(\mathbf{I}_{I} - \mathbf{g}_{t}\mathbf{x}_{t}^{\top}\right)\boldsymbol{\psi}_{t-1}
		 + \langle \beta \rangle_{t}\mathbf{S}_{t}\mathbf{x}_{t}
	  	 \left(y_{t} - \mathbf{x}_{t}^{\top}\boldsymbol{\mu}_{t-1}^{\mathbf{w}} - \langle \mu \rangle_{t}\right).
\end{split}
\end{equation}
For the recursion to compute $\mathbf{S}$, we first rewrite the weight covariance matrix as follows:
\begin{equation}
	\boldsymbol{\Sigma}_{t}^{\mathbf{w}}
	= \mathbf{P}_{t-1}^{\mathbf{w}} - \frac{\mathbf{P}_{t-1}^{\mathbf{w}}\mathbf{x}_{t}\mathbf{x}_{t}^{\top}\mathbf{P}_{t-1}^{\mathbf{w}}}
	  {\mathbf{x}_{t}^{\top}\mathbf{P}_{t-1}^{\mathbf{w}}\mathbf{x}_{t} + \langle \beta \rangle_{t}^{-1}}.
\end{equation}
Thus,
\begin{equation}
\begin{split}
	\mathbf{S}_{t}
	&= \mathbf{S}_{t-1} 
	  - \frac{\mathbf{S}_{t-1}\mathbf{x}_{t}\mathbf{x}_{t}^{\top}\mathbf{P}_{t-1}^{\mathbf{w}} 
	          + \mathbf{P}_{t-1}^{\mathbf{w}}\mathbf{x}_{t}\mathbf{x}_{t}^{\top}\mathbf{S}_{t-1}}
	         {\mathbf{x}_{t}^{\top}\mathbf{P}_{t-1}^{\mathbf{w}}\mathbf{x}_{t} + \langle \beta \rangle_{t}^{-1}}
      + \frac{\mathbf{P}_{t-1}^{\mathbf{w}}\mathbf{x}_{t}\left(\mathbf{x}_{t}^{\top}\mathbf{S}_{t-1}\mathbf{x}_{t}\right)
   		      \mathbf{x}_{t}^{\top}\mathbf{P}_{t-1}^{\mathbf{w}}}
	     	 {\left(\mathbf{x}_{t}^{\top}\mathbf{P}_{t-1}^{\mathbf{w}}\mathbf{x}_{t} + \langle \beta \rangle_{t}^{-1}\right)^{2}}	\\
	&= \mathbf{S}_{t-1} - \mathbf{S}_{t-1}\mathbf{x}_{t}\mathbf{g}_{t}^{\top} - \mathbf{g}_{t}\mathbf{x}_{t}^{\top}\mathbf{S}_{t-1}
	   + \mathbf{g}_{t}\mathbf{x}_{t}^{\top}\mathbf{S}_{t-1}\mathbf{x}_{t}\mathbf{g}_{t}^{\top}	\\
	&= \left(\mathbf{I}_{I} - \mathbf{g}_{t}\mathbf{x}_{t}^{\top}\right)\mathbf{S}_{t-1}\left(\mathbf{I}_{I} - \mathbf{x}_{t}\mathbf{g}_{t}^{\top}\right).
\end{split}
\end{equation}

\section*{Applications}

\subsection*{Practicalities}

Here, we show that the measurement-noise density $p(\eta_{t}|\epsilon)$ of the BYPASS framework naturally induces a symmetric Beta prior of the second kind \cite{johnson95} over the insensitivity hyperparameter $\epsilon$. First of all, we note that this density can be written as a mixture of a Uniform distribution and a truncated Laplace distribution:
\begin{equation}
\begin{split}
	p\left(\eta^{y}_{t}|\epsilon\right)
	&= \frac{1}{2(1+\epsilon)}
		 \bigg\{\mathds{1}_{\left[-\epsilon, \epsilon\right]}\left(\eta^{y}_{t}\right)
						+ \left[1 - \mathds{1}_{\left[-\epsilon, \epsilon\right]}\left(\eta^{y}_{t}\right)\right]
							e^{\epsilon - \left|\eta^{y}_{t}\right|}\bigg\} \\
	&= \frac{\epsilon}{1+\epsilon}\mathcal{U}\left(\eta^{y}_{t}|-\epsilon, \epsilon\right) 
		 + \frac{1}{1+\epsilon}\mathcal{L}_{\left[-\epsilon, \epsilon\right]}^{\text{trunc}}\left(\eta^{y}_{t}|0, 1\right).
\end{split}
\end{equation}
Next, let $\pi \equiv \epsilon/(1+\epsilon)$ be the mixing coefficient in the above mixture. A natural (and widely used) prior over $\pi$ is a symmetric Beta\footnote{This distribution is equivalent to a symmetric Dirichlet distribution with 2 states.}:
\begin{equation}
\label{eq:mixingprior}
	p\left(\pi\right)
	= \mathcal{B}\left(\pi|s\right)
	= \frac{1}{B\left(s, s\right)}\left[\pi\left(1 - \pi\right)\right]^{s - 1},
\end{equation}
where $B(\cdot, \cdot)$ is the Beta function and $s> 0$ denotes the shape of the distribution. Since $\epsilon = \pi/(1 - \pi)$ by definition, it immediately follows that $\epsilon$ has a symmetric Beta distribution of the second kind \cite{johnson95}. That is, for $\epsilon > 0$,
\begin{equation}
	p\left(\epsilon\right)
	= \frac{1}{B\left(s, s\right)}\epsilon^{s - 1}\left(1 + \epsilon\right)^{-2s}.
\end{equation}
The mean of this distribution is given by $s / (s-1)$.

\subsubsection*{References}

\renewcommand\refname{\vskip -0.9cm}

\small
{
\bibliographystyle{unsrt}
\bibliography{bypass}
}
